\documentclass[10pt, a4paper]{article}

\usepackage{lrec2026} 
\usepackage{float} 
\usepackage{placeins}

\usepackage{amsmath}
\usepackage{graphicx}
\usepackage{subcaption}
\usepackage{xcolor}
\usepackage{soul} 
\usepackage{todonotes}
\usepackage{csquotes} 
\usepackage{multirow}
\usepackage{booktabs}
\usepackage{adjustbox}
\usepackage{tabularx}
\usepackage{makecell}
\usepackage{array}
\usepackage{textcomp}

\usepackage{iftex}
\ifXeTeX\else
  \errmessage{This paper must be compiled with XeLaTeX}
\fi

\usepackage{fontspec}
\usepackage{polyglossia}

\setdefaultlanguage{english}
\setotherlanguages{greek,ukrainian,hebrew,georgian,arabic,persian,chinese}

\defaultfontfeatures{Ligatures=TeX}

\newfontfamily\greekfont[Path=fonts/,Script=Greek]{NotoSerif-Regular.ttf}
\newfontfamily\ukrainianfont[Path=fonts/,Script=Cyrillic]{NotoSerif-Regular.ttf}

\newfontfamily\hebrewfont[Path=fonts/,Script=Hebrew]{NotoSansHebrew-Regular.ttf}
\newfontfamily\georgianfont[Path=fonts/,Script=Georgian]{DejaVuSans.ttf}

\newfontfamily\arabicfont[Path=fonts/,Script=Arabic]{NotoNaskhArabic-Regular.ttf}
\newfontfamily\persianfont[Path=fonts/,Script=Arabic,Language=Persian]{NotoNaskhArabic-Regular.ttf}

\newfontfamily\chinesefont{FandolSong-Regular.otf}

\providecommand{\textgeorgian}[1]{{\georgianfont #1}}
\providecommand{\texthebrew}[1]{{\hebrewfont #1}}
\providecommand{\textarabic}[1]{{\arabicfont #1}}
\providecommand{\textpersian}[1]{{\persianfont #1}}

\providecommand{\textgreek}[1]{{\greekfont #1}}
\providecommand{\textchinese}[1]{{\chinesefont #1}}

\newcommand{\numLanguages}{34} 
\newcommand{\numPIEtxt}{3054}
\newcommand{\numPIEimg}{7040}
\title{A Parallel Cross-Lingual Benchmark for Multimodal Idiomaticity Understanding}

\name{
\parbox{\linewidth}{\centering
Dilara Torunoğlu-Selamet, Dogukan Arslan, Rodrigo Wilkens, Wei He, 
Doruk Eryiğit, Thomas Pickard, Adriana S. Pagano, Aline Villavicencio, Gülşen Eryiğit, Ágnes Abuczki, Aida Cardoso, Alesia Lazarenka, Dina Almassova, Amalia Mendes, Anna Kanellopoulou, Antoni Brosa-Rodríguez, Baiba Saulite, Beata Wojtowicz, Bolette Pedersen, Carlos Manuel Hidalgo-Ternero, Chaya Liebeskind, Danka Jokić, Diego Alves, Eleni Triantafyllidi, Erik Velldal, Fred Philippy, Giedre Valunaite Oleskeviciene, Ieva Rizgeliene, Inguna Skadina, Irina Lobzhanidze, Isabell Stinessen Haugen, Jauza Akbar Krito, Jelena M. Marković, Johanna Monti, Josue Alejandro Sauca, Kaja Dobrovoljc, Kingsley O. Ugwuanyi, Laura Rituma, Lilja Øvrelid, Maha Tufail Agro, Manzura Abjalova, Maria Chatzigrigoriou, María del Mar Sánchez Ramos, Marija Pendevska, Masoumeh Seyyedrezaei, Mehrnoush Shamsfard, Momina Ahsan, Muhammad Ahsan Riaz Khan, Nathalie Carmen Hau Norman, Nilay Erdem Ayyıldız, Nina Hosseini-Kivanani, Noémi Ligeti-Nagy, Numaan Naeem, Olha Kanishcheva, Olha Yatsyshyna, Daniil Orel, Petra Giommarelli, Petya Osenova, Radovan Garabik, Regina E. Semou, Rozane Rebechi, Salsabila Zahirah Pranida, Samia Touileb, Sanni Nimb, Sarfraz Ahmad, Sarvinoz Sharipova, Shahar Golan, Shaoxiong Ji, Sopuruchi Christian Aboh, Srdjan Sucur, Stella Markantonatou, Sussi Olsen, Vahide Tajalli, Veronika Lipp, Voula Giouli, Yelda Yeşildal Eraydın, Zahra Saaberi and Zhuohan Xie
}
}

\address{}

\abstract{
Potentially idiomatic expressions (PIEs)  construe meanings inherently tied to the everyday experience of a given language community. As such, they constitute an interesting challenge 
for assessing the linguistic (and to some extent cultural) capabilities of NLP systems. In this paper, we present XMPIE, a parallel multilingual and multimodal dataset of potentially idiomatic expressions. The dataset, containing  \numLanguages{} languages and over ten thousand items, allows comparative analyses of idiomatic patterns among language-specific realisations and preferences in order to gather insights about shared cultural aspects. This parallel dataset allows to evaluate model performance for a given PIE in different languages and whether idiomatic understanding in one language can be transferred to another. Moreover, the dataset supports the study of PIEs across textual and visual modalities, to measure to what extent PIE understanding in one modality transfers or implies in understanding in another modality (text vs. image). The data was created by language experts, with both textual and visual components crafted under multilingual guidelines, and 
each PIE is accompanied by five images representing a spectrum from idiomatic to literal meanings, including semantically related and random distractors. The result is a high-quality benchmark for evaluating multilingual and multimodal idiomatic language understanding.\\ \newline
\Keywords{Multiword Expressions, Machine Translation, Multilingual models, Multimodal models}}

\begin{document}

\maketitleabstract

\section{Introduction}


As a widely studied class of non-compositional multiword expressions, idioms such as ``green fingers'' and ``kick the bucket'', pose persistent challenges for both humans and natural language processing (NLP) systems \cite{10.1007/3-540-45715-1_1}. Due to the cultural knowledge and shared conceptualisations they embody, idioms can be challenging not only for (non-native) speakers
\citep{charteris2002second,kovecses2010metaphor}, but also in 
tasks like machine translation
\citep{10.1007/3-540-45715-1_1}.  
Indeed, idioms bring to light a range of cross-linguistic dynamics. On the one hand, speakers of closely related languages often share similar idiomatic realisations, while on the other hand,  idioms may be specific to a linguistic community and lexically non-transferable and opaque, leading to entirely different figurative mappings \citep{irujo1986don}. For example, ``bad apple'' is directly transferable into Turkish as ``çürük elma'' (lit. ``rotten apple''), whereas the English idiom ``bear market''  has no idiomatic equivalent and is instead paraphrased descriptively (e.g., ``düşen piyasa'' lit. ``declining market''), 
while the Brazilian Portuguese ``levar uma bola nas costas'' (lit. ``take a ball on the back'') conveys the idea of being betrayed 
may be understood by a Turkish speaker 
given the analogous ``sırtından bıçaklanmak'' (``to take a knife in the back''), with a similar metaphorical frame of unexpected betrayal. 
This cultural and cross-lingual dimension positions idioms as an ideal and important testbed for investigating both linguistic diversity and the capacity of NLP models to generalize underlying meaning across languages.


\begin{figure*}[!htp]
\centering

\begin{subfigure}{0.10\textwidth}
\centering
\small{(a) idiomatic}
\includegraphics[width=\linewidth]{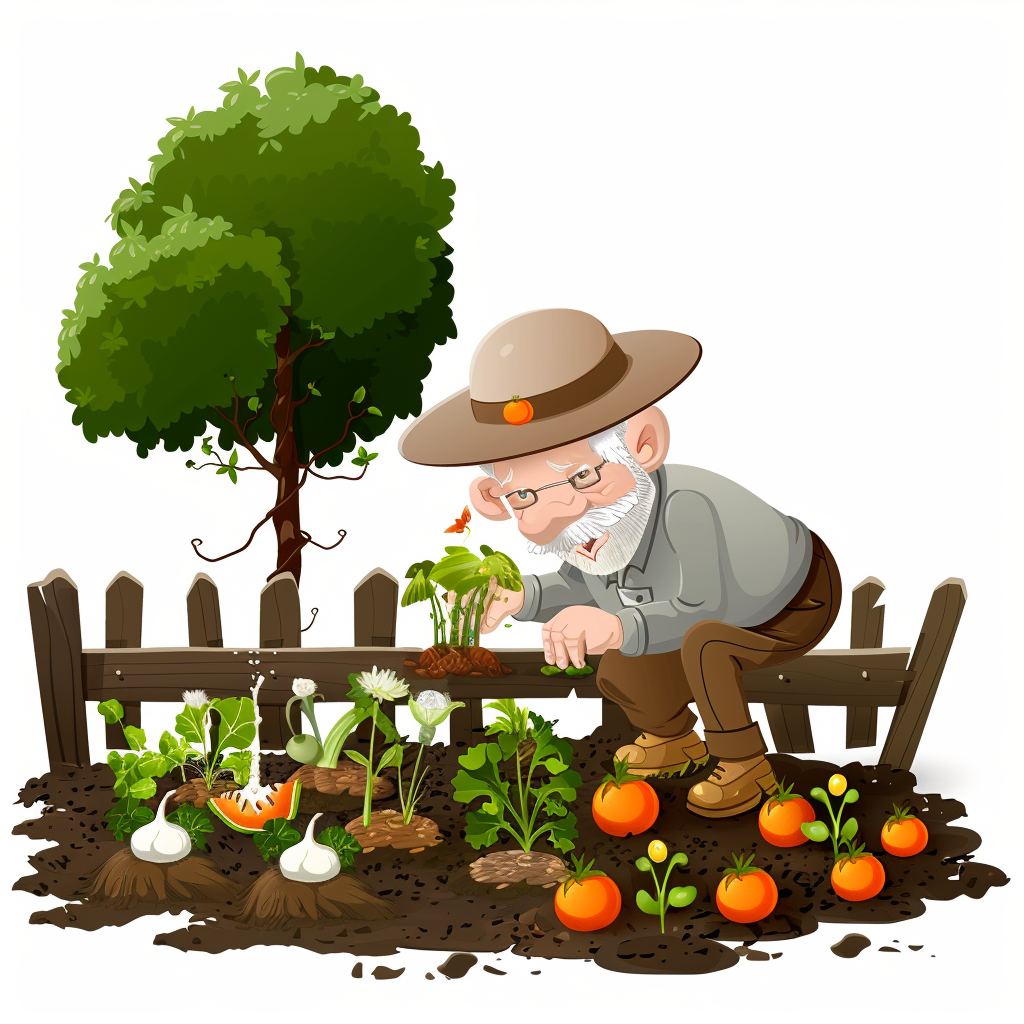} 
\label{fig:subim1}
\end{subfigure}
\hfill
\begin{subfigure}{0.10\textwidth}
\centering
\small{(b) idiomatic-related}
\includegraphics[width=\linewidth]{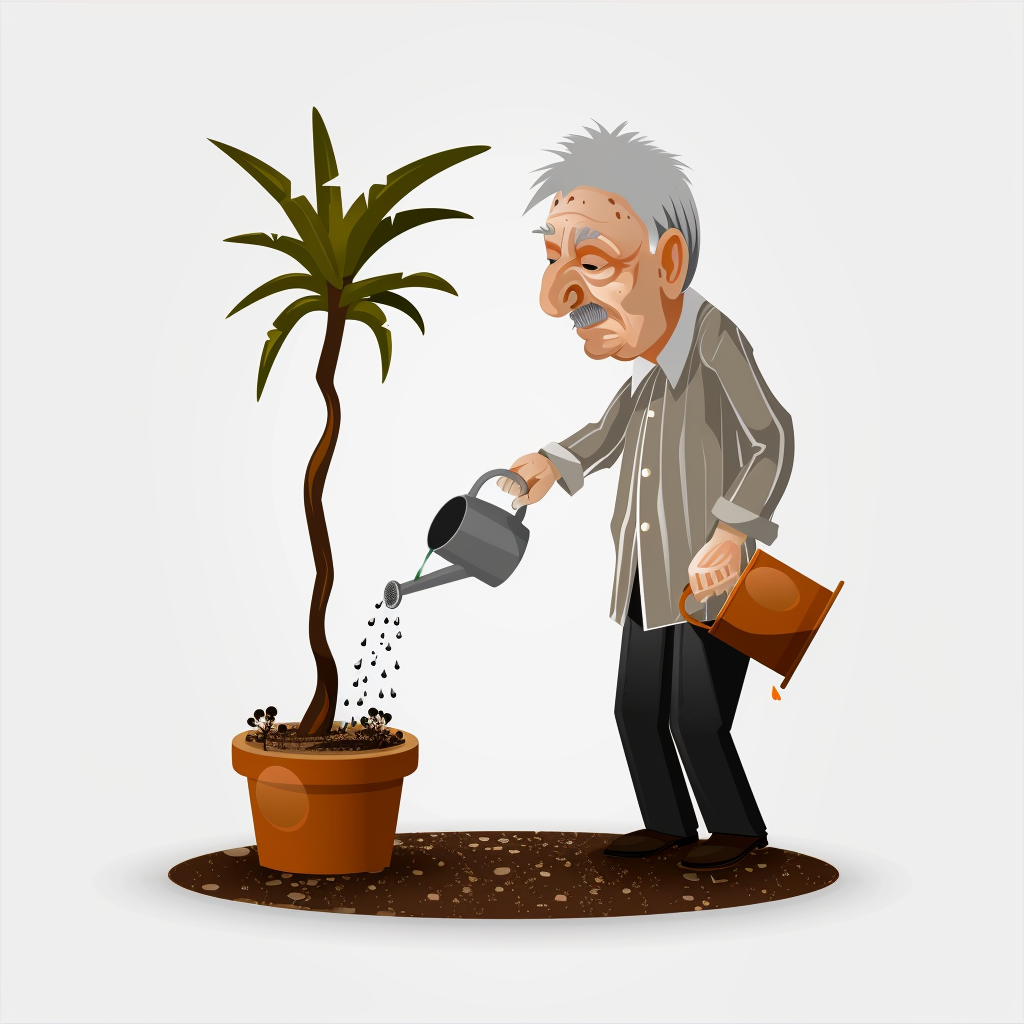}
\label{fig:subim2}
\end{subfigure}
\hfill
\begin{subfigure}{0.10\textwidth}
\centering
\small{(c) literal-related}
\includegraphics[width=\linewidth]{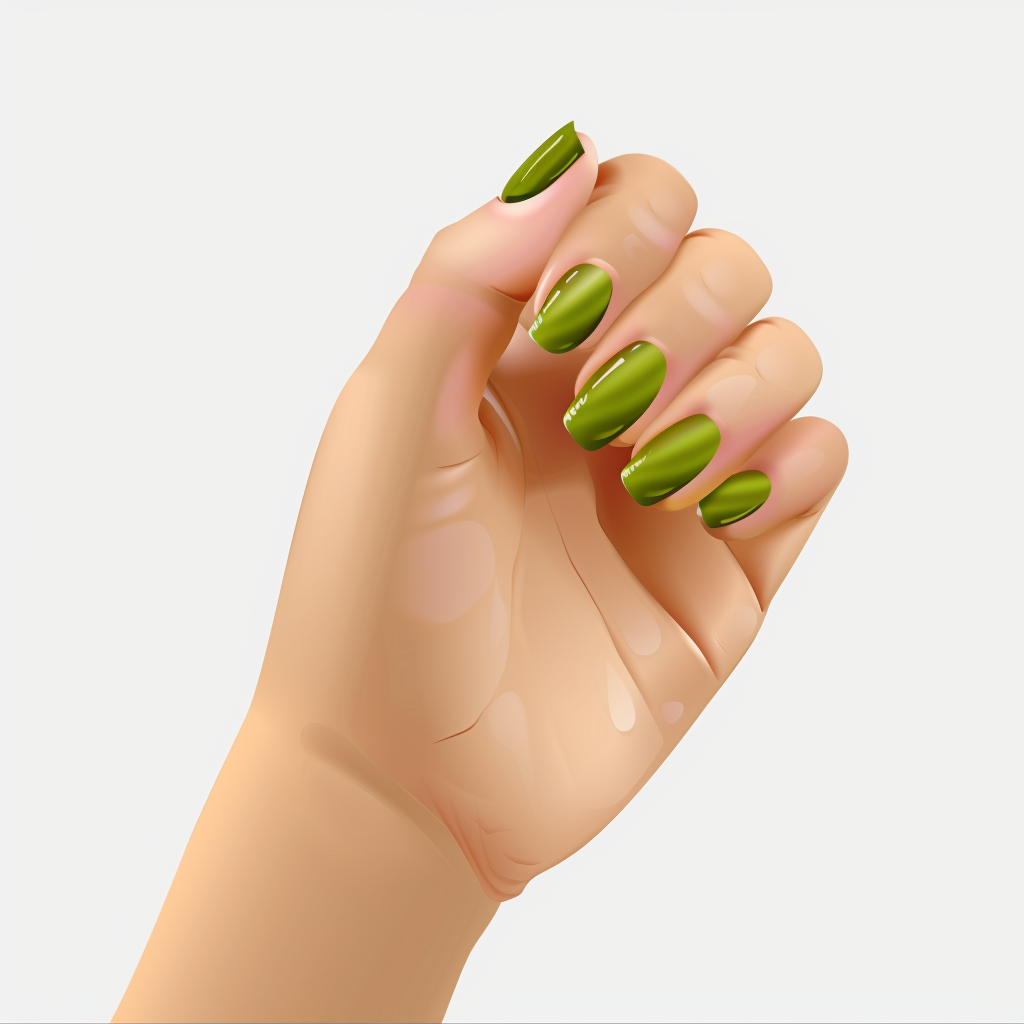}
\label{fig:subim3}
\end{subfigure}
\hfill
\begin{subfigure}{0.10\textwidth}
\centering
\small{(d) literal}
\includegraphics[width=\linewidth]{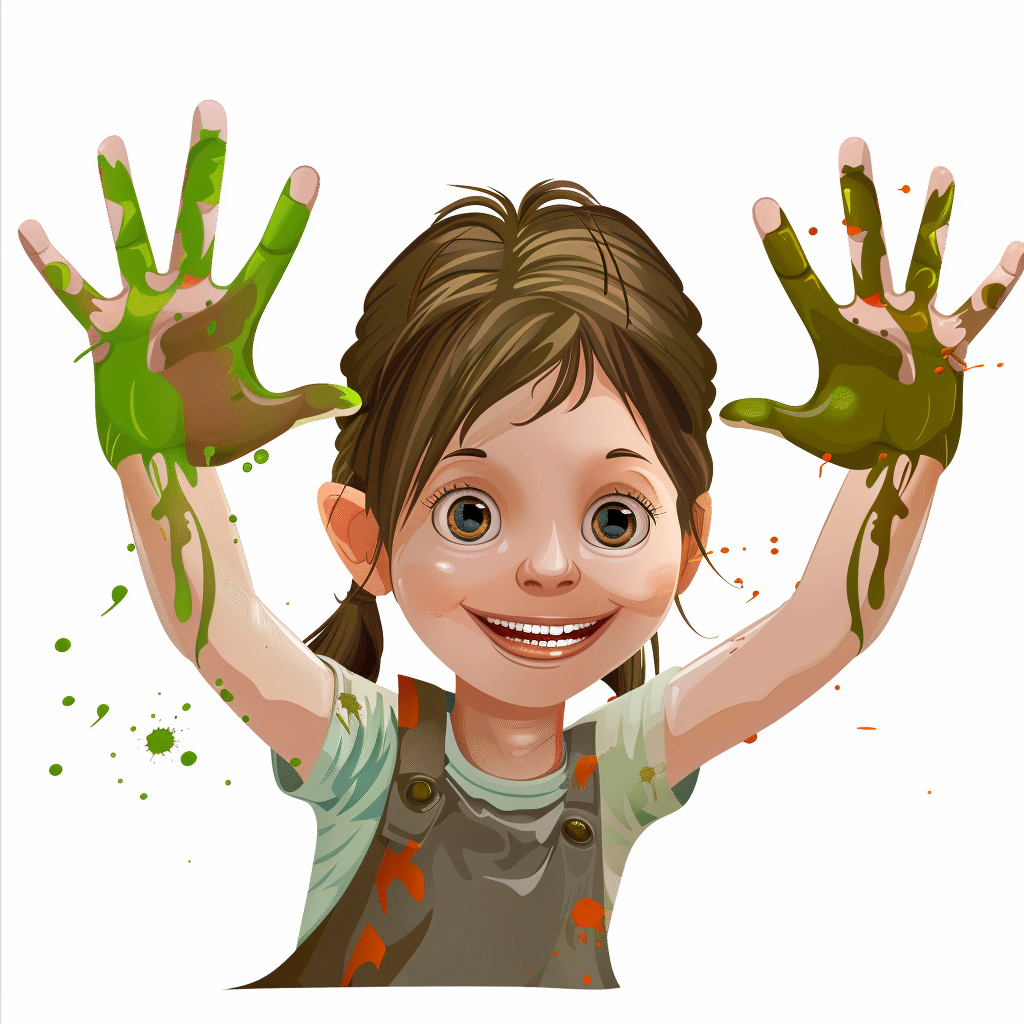}
\label{fig:subim4}
\end{subfigure}
\hfill
\begin{subfigure}{0.10\textwidth}
\centering
\small{(e) random distractor}
\includegraphics[width=\linewidth]{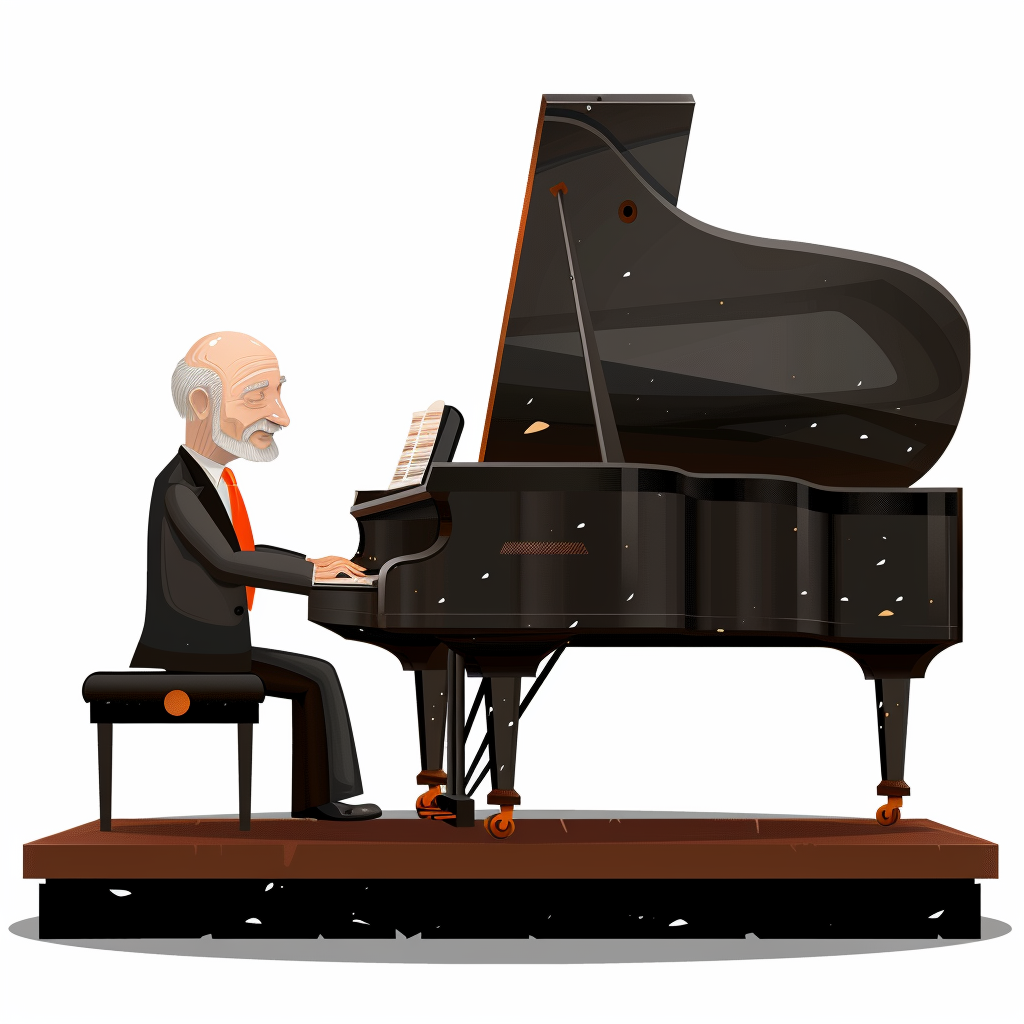}
\label{fig:subim5}
\end{subfigure}

\raggedright{
\footnotesize{


}
}


\caption{Image data for \textit{``green fingers''} idiomatically denoting  \emph{skill in gardening.}}

\label{fig:image2}
\end{figure*}

There has been a growing emphasis on cultural benchmarks \citeplanguageresource{chiu-etal-2025-culturalbench,khanuja-etal-2024-image,romero2024cvqa} when evaluating large language models (LLMs), reflecting the need to assess, not only their general linguistic competence, but also their sensitivity to diverse cultural contexts and knowledge. 
Since idioms are deeply embedded in cultural knowledge and often resist literal translation, the introduction of a multimodal parallel dataset of potentially idiomatic expressions (PIEs) provides significant value for facilitating crosslingual intercultural comparison and systematically evaluating multilingual representativeness in LLMs.
Moreover, PIEs are an interesting challenge 
 due to the data-intensive nature of LLMs, which creates a serious obstacle to building accurate representation for low-frequency and long-tail phenomena, like PIEs, especially when considering multilinguality and lower resource  scenarios.  

While contemporary LLMs achieve strong performance on a variety of NLP tasks \citep{zubiaga2024nlp}, they often struggle with idiomatic expressions, highlighting a gap in their capacity for idiomatic understanding \citep{phelps-etal-2024-sign,arslan-etal-2025-using}. Traditional idiom processing benchmarks, which typically frame idiom comprehension as a classification task, have been criticized as not fully reflective of a model's grasp of idiomatic meaning \citep{boisson-etal-2023-construction,he-etal-2025-idiomaticity}, particularly given that general classification performance of LLMs is still limited. 
As alternatives, paraphrasing and multimodal approaches have been proposed to better assess idiomatic competence, with recent datasets integrating visual modalities alongside text to challenge models in multimodal idiom understanding \citeplanguageresource{pickard-etal-2025-semeval}.




This paper introduces a benchmark that paves the way for the evaluation of the extent to which LLMs display accurate  
idiom comprehension across languages and across modalities. 
The \textit{Cross-lingual and Multimodal Potentially Idiomatic Expressions} (XMPIE) dataset, a parallel cross-lingual benchmark for multimodal idiomaticity understanding. XMPIE is inspired by AdMIRe 1.0 \citeplanguageresource{pickard-etal-2025-semeval} which covers 2 languages (English and Portuguese),  and extends it to \numLanguages{} languages, ranging from widely spoken ones to those endangered either sociolinguistically (viz., Aromanian and Luxembourgish) or digitally (viz., Igbo, Catalan, Greek, Latvian, and Lithuanian) according to UNESCO \citep{Moseley2010} and the European Language Equality (ELE) report \citep{ananiadou2012english}. 
Starting from seed English PIEs and extending them to these languages 
XMPIE has 
\numPIEtxt{} PIEs and 
\numPIEimg{} in total for figurative and literal meanings, along with distractors (Figure~\ref{fig:image2}).
In the paper, we discuss the construction of this parallel cross-linguistic dataset for understanding multimodal idiomaticity and the analyses we carried out on this benchmark. The paper starts with related work in $\S$\ref{sec:relatedwork}, the  annotation methodology in $\S$\ref{sec:annotation}
$\S$\ref{sec:analysis} presents the resulting dataset and $\S$\ref{sec:results} the benchmark evaluation, and $\S$\ref{sec:conclusion} the conclusions.



\section{Related Work}
\label{sec:relatedwork}

\begin{table*}[htp!]
\footnotesize
\centering
\begin{tabular}{@{}llll@{}}
\toprule

\textbf{Dataset} & \textbf{\#Size} & \textbf{\#Idioms} & \textbf{Language}                             \\ \midrule

VNC-Tokens \small{\citeplanguageresource{cook2008vnc}}                & 2,566   & 53    & en  \\
Open-MWE \small{\citeplanguageresource{Hashimoto2009}}                & 102,856 & 146   & ja  \\
Sporleder and Li \small{\citeplanguageresource{sporleder-li-2009-unsupervised}}   & 3,964   & 17    & en  \\
IDIX \small{\citeplanguageresource{sporleder-etal-2010-idioms}}                  & 5,836   & 78    & en \\
SemEval-2013 Task 5b \small{\citeplanguageresource{korkontzelos-etal-2013-semeval}}   & 4,350   & 65    & en \\
\multirow{4}{*}{PARSEME \small{\citeplanguageresource{savary:hal-01223349}}}         & \multirow{4}{*}{274,376}           & \multirow{4}{*}{13,755}          & \multirow{4}{2cm}{bg, cs, fr, de, he, it, lt, mt, el, pl, pt, ro, sl, es, sv, tr} \\ \\ \\ \\
MAGPIE  \small{\citeplanguageresource{haagsma-etal-2020-magpie}}                 & 56,622  & 2,007 & en \\
EPIE  \small{\citeplanguageresource{https://doi.org/10.48550/arxiv.2006.09479}}                   & 25,206  & 717   & en             \\
AStitchInLang.Models \small{\citeplanguageresource{tayyar-madabushi-etal-2021-astitchinlanguagemodels-dataset}} & 6,430   & 336   & en, pt         \\
ID10M$_{silver}$ \small{\citeplanguageresource{Tedeschi2022}}            & 800     & 470   & de, en, es, it \\
\multirow{3}{*}{ID10M$_{gold}$ \small{\citeplanguageresource{Tedeschi2022}}}      & \multirow{3}{*}{262,781}            & \multirow{3}{*}{10,118}          & \multirow{3}{2cm}{de, en, es, fr, it, ja, nl, pl, pt, zh} \\                         \\ \\
SemEval-2022 Task 2 \small{\citeplanguageresource{TayyarMadabushi2022}}      & 8,683   & 50    & en, gl, pt     \\
Dodiom \small{\citeplanguageresource{Eryiit2022}}                   & 12,706  & 73    & it, tr  \\    
idiom-corpus-llm \small{\citeplanguageresource{arslan-etal-2025-using}}                   & 34,600  & 173    & en, ja, it, tr   \\ \bottomrule
\end{tabular}%
\caption{A summary of various idiom corpora, detailing the number of sentences, the number of idioms, and the languages included for each \protect\citeplanguageresource{arslan-etal-2025-using}.}
\label{tab:dataset}
\end{table*}


Understanding how computational models represent and predict compositional meaning relies notably on the development of  benchmark datasets. These resources have advanced, from early, monolingual collections with compositionality ratings  \citeplanguageresource{cook2008vnc} to more complex multilingual corpora that capture nuanced, context-dependent phenomena \citeplanguageresource{haagsma-etal-2020-magpie}. In parallel with advancements in human-annotated data, recent work has also begun to explore innovative strategies for corpus creation and the integration of multimodal evaluation items. 

A summary of these datasets is shown in Table~\ref{tab:dataset}, where a prevalence of European languages can be seen. 
%
Despite  advances in scale and diversity, the majority of existing idiom datasets are not constructed in a parallel fashion, meaning that even in multilingual datasets, idiomatic expressions are not systematically aligned across languages. As a result, while these corpora enable monolingual and, to some extent, multilingual evaluation, they do not facilitate direct cross-linguistic comparison of idiom usage or support fine-grained investigations of how idiomatic meaning is preserved or altered across languages. This lack of parallelism remains a critical gap, particularly for research on multilingual representation learning and cross-lingual transfer.




Foundational work in this area was primarily monolingual: from 
\citetlanguageresource{reddy-etal-2011-empirical} English noun compound dataset, which established a strong link between literal word meaning and overall phrase compositionality, and were  extended by  \citetlanguageresource{10.1162/coli_a_00341} with human compositionality judgments for nominal compounds in English, French, and Portuguese, supporting multilingual evaluation.  
Other multilingual resources include SemEval-2022 Task 2 \citeplanguageresource{tayyar-madabushi-etal-2022-semeval,tayyar-madabushi-etal-2021-astitchinlanguagemodels-dataset} 
providing 8,683 multilingual entries for idiomaticity detection and sentence representation for English, Portuguese and Galician. 
MultiCoPIE \citeplanguageresource{sentsova-etal-2025-multicopie} is a multilingual corpus of PIEs in Catalan, Italian, and Russian, with 
annotations specifically designed to analyze factors like lexical overlap in cross-lingual transfer learning.

Beyond increasing language coverage, research has also focused on capturing more granular, context-dependent aspects of idiomaticity. For instance, \citetlanguageresource{he-etal-2025-idiomaticity} introduced a large-scale dataset for English and Portuguese containing minimal pairs, human judgments at both type and token levels, paraphrases, and contextual occurrences. Another example is DICE \citeplanguageresource{mi-etal-2025-rolling}, a contrastive dataset designed specifically to test whether large language models can effectively use context to disambiguate idiomatic expressions and to systematically analyze their limitations.

A more recent and challenging frontier is the extension of this research into multimodal settings, exploring how idiomatic meaning is conveyed through both text and images. The Image Recognition of Figurative Language (IRFL) dataset from \citetlanguageresource{yosef-etal-2023-irfl}, which established that state-of-the-art vision-language models significantly underperform humans on multimodal tasks involving metaphors, similes, and idioms. Building on this, \citetlanguageresource{saakyan-etal-2025-understanding} developed V-FLUTE, a dataset that adds a layer of explainability by requiring models to generate textual justifications for their visual entailment decisions on figurative language. The scope of multimodal research has also expanded to specific languages and domains. For instance, \citetlanguageresource{10.1609/aaai.v39i24.34728} created MChIRC, a large-scale dataset specifically for Chinese idiom comprehension, while \citetlanguageresource{tong2025hummusdatasethumorousmultimodal} compiled the HUMMUS dataset to analyze the interplay of humorous metaphor and idioms.

Parallel to the development of these annotated corpora, recent studies have explored innovative strategies for data creation to overcome the costs of manual annotation. One direction involves novel human-in-the-loop methods, such as the gamified crowdsourcing framework introduced by \citetlanguageresource{Eryiit2022} 
to engage native speakers in creating and rating idiom examples. In a complementary, model-centric approach, \citetlanguageresource{arslan-etal-2025-using} investigated using LLMs themselves to generate synthetic idiom corpora, offering a potentially more scalable and efficient alternative to human-annotated datasets.

However, a massively cross-lingually aligned resource for exploring idiomaticity in both language production and model evaluation is still missing. Therefore,
in this paper, we address this and present a parallel dataset across multiple languages and modalities and  discuss the protocol adopted  
to facilitate annotation and coordination.

\section{Annotation Methodology}
\label{sec:annotation}


Starting from a seed set of PIEs in English, 89 native or highly fluent language experts collaboratively
identified equivalent idioms in their respective language variants
 and generated
relevant images depicting both figurative and literal
meanings along with distractors using image generation systems.

Language experts were recruited through open invitation through UNIDIVE \cite{savary2024unidive} and participated in three online workshops, where they were presented with step-by-step instructions about the annotation process. They were also given  
written guidelines 
while additional consultation was offered upon demand.

%
For each seed English PIE (e.g., ``bad apple''), annotators provided: (1) a literal word-by-word translation into their language (e.g., Ukrainian ``gnile âbluko''); (2) a transliterated version of the literal translation (when applicable) (``\foreignlanguage{ukrainian}{гниле яблуко}''); (3)~the idiomatic equivalent of the PIE  
in their language (``paršiva vìvcâ''); (4) its literal (word-by-word) English translation\footnote{We included literal translation of the idiomatic equivalent back into English for facilitating analyses and enabling cross-linguistic semantic transparency.} (``lousy sheep'')  and (5) a transliterated version of that idiomatic form (when applicable) (``\foreignlanguage{ukrainian}{паршива вівця}'').

The second step was to generate images. To do this, we used Discord\footnote{https://discord.com/}
to facilitate communication and sharing during the annotation process from a centralized point and to enable automatic collection of image generation prompts. 
Each language was assigned its own Discord channel for the language experts to collaborate in preparing the PIE images and to conduct the necessary discussions. 
Midjourney\footnote{https://www.midjourney.com/} was adopted 
for image generation 
and each language was provided with a one-month Midjourney subscription. 
The administrators were members of all channels, monitored the process, and responded to annotators' questions as needed.

For PIE image generation, annotators produced \emph{five} candidate images that range from the idiomatic to the  literal meaning (Figure~\ref{fig:image2}): (1) an image of the  \textit{idiomatic} sense, (2) an image for an \textit{idiomatic–related} sense, (3)~an image with a  \textit{literal–related} sense, (4) one for the \textit{literal} sense, and (5) one for an  \textit{unrelated} random distractor. 
This provides an image representation for the idiomatic and literal meanings of a PIE as well as easy and difficult distractors. 
Textual prompts providing descriptions for each of the five target senses were used to generate images. 
Language experts identified the two target senses (idiomatic and literal) before generating the three distractors.
Where possible, the semantically related distractors stayed in the same broad semantic category as the target (e.g., an object for object-denoting PIE). 
For generating the semantically related  distractors, strategies included focusing on the individual words or on aspects of the specific senses
(e.g. ``a bag of apples'' for ``bad apple'' and ``a shelf with small pieces of cheese'' for ``big cheese''). 
One of the challenges is that prompts need to be framed in terms of concrete and visually grounded descriptions (``recipe book'' rather than the more abstract ``instructions'') and avoiding potential issues that models struggle with (e.g., legible text on signs; fine-grained hand poses; subtle emotions). Abstract adjectives were conveyed via visual proxies (e.g., for an important person the prompt would include a figure of authority). When a figurative sense is hard to portray directly, a literal-looking scene that cues the idiomatic reading (e.g., a reviewer or judge for ``armchair critic'') would be used. During training, ``good'' and ``bad'' prompt examples were provided to the annotators for calibration. 
Image prompts were executed in private (``stealth'') mode via direct messages with the Midjourney bot to prevent public leakage of the images.
Final images were saved as high-resolution PNGs named 1–5 for idiomatic, idiomatic-related, literal-related, literal, and random distractor types.





\section{The XMPIE Dataset}
\label{sec:analysis}

\begin{figure}[htp!]
    \includegraphics[width=\linewidth]{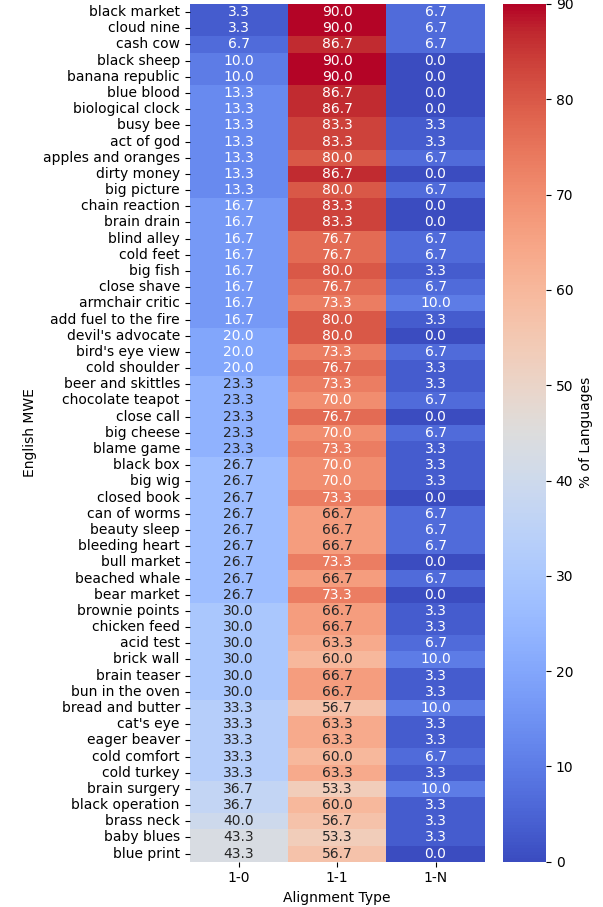}
    \caption{Percentage distribution of alignment types (\textit{1-0}, \textit{1-1}, \textit{1-N}) across PIEs. Only idioms in more than 20 languages shown.}
    \label{fig:translation_by_mwe}
\end{figure}


\begin{table*}[htbp!]
\centering
\renewcommand{\arraystretch}{1.3}
\begin{adjustbox}{max width=\textwidth}
\begin{tabular}{p{1.2cm} 
p{4.2cm} p{11.8cm}}
\toprule
\textbf{Alignment Type} & 
\textbf{Nature of Mapping} & \textbf{Examples} \\
\midrule

1–1 & 
Target idiom has identical or near-identical wording (lexis and/or syntactical structure) and meaning &
\small
\underline{‘bad apple’} $\rightarrow$ çürük elma (‘rotten apple’, Turkish); \textgeorgian{ცუდი ნაყოფი} ‘cudi naqop’i’ (‘bad fruit’, Georgian); \texthebrew{תפוח רקוב} (‘rotten apple’, Hebrew); maçã podre (‘apple rotten’, Brazilian Portuguese). 
\underline{‘add fuel to the fire’} $\rightarrow$ yangına körükle gitmek (‘to go to the fire with a bellows’, Turkish); hælde benzin på båлет (‘pour gasoline to the fire’, Danish); \textgreek{ρίχνω λάδι στη φωτιά} (‘throw oil on the fire’, Greek); deitar achas na fogueira (‘lay down woodsticks on the bonfire’, Portuguese); colocar lenha na fogueira (‘put woodsticks in the bonfire’, Brazilian Portuguese); įpilti aliejaus į ugnį (‘pour oil into the fire’, Lithuanian); afegir llenya al foc (‘add firewood to the fire’, Catalan). 
\underline{‘apples and oranges’} $\rightarrow$ “epler og pærer” (‘apples and pears’, Norwegian); Äppel a Bieren (‘apples and pears’, Luxembourgish); jabolka in hruške (‘apples and pears’, Slovenian).
\underline{‘big fish’} $\rightarrow$ \textgeorgian{დიდი მოთამაშე} ‘didi motamashe’ (‘big player’, Georgian).\\[3pt]

& 
Target idiom has different wording (lexis and/or syntactical structure) but same conceptual meaning &
\small
\underline{‘bad apple’} $\rightarrow$ развалена стока (‘rotten goods’, Bulgarian). 
\underline{‘eager beaver’} $\rightarrow$ \textarabic{الله تان بیر گون اوغورلامیش}/Allahtan bir gün oğurlamış (‘stealing one day from God’, Azeri). 
\underline{‘ancient history’} $\rightarrow$ \textarabic{کهنه پالان}/ köhnə palan (‘old packsaddle’, Azeri); acqua passata (‘water passed’, Italian). 
\underline{‘beer and skittles’} $\rightarrow$ \textarabic{تزه کوزه سرین سو}/təze küzə sərin su (‘new clay jug, cold water’, Azeri). 
\underline{‘add fuel to the fire’} $\rightarrow$\textpersian{آتش بیار معرکه شدن}/atash biyar-e mareke shodan (‘fire-bringer of the battlefield’, Farsi).\\\hline

1–0 & 
Target language uses descriptive (non-idiomatic) expression or has no equivalent &
\small
\underline{‘chocolate teapot’} $\rightarrow$ \textgeorgian{აბსოლუტურად უსარგებლო რამ} (‘absolutely useless thing’, Georgian). 
\underline{‘brain drain’} $\rightarrow$\textpersian{فرار مغزها}/farar-e maghzhā (‘escape of brains’, Farsi). 
\underline{‘agony aunt’} $\rightarrow$ conselheiro sentimental (‘sentimental counsellor’, Portuguese); \textgeorgian{მკითხველის მრჩეველი} (‘reader’s advisor’, Georgian). 
\underline{‘best man’} $\rightarrow$ apoio do noivo (‘support of the groom’, Portuguese); padrinho de casamento (‘wedding godfather’, Portuguese). 
\\[3pt]

& 
Target language uses idiom made up of single lexical item (not an MWE) &
\small
\underline{‘brain teaser’} $\rightarrow$ pazli (Georgian); nøtt (‘nut’, Norwegian). 
\underline{‘busy bee’} $\rightarrow$ futkari (Georgian). 
\underline{‘brain teaser’} $\rightarrow$ \textarabic{تاپپاجا}/tappaca (Azeri);\textpersian{معما}/moamma (Farsi). 
\underline{‘cheat sheet’} $\rightarrow$ cola (Portuguese); skonaki (Greek); пищов (‘pishtov’, Bulgarian); puska (Hungarian). 
\underline{‘chicken feed’} $\rightarrow$ drobtinice (‘breadcrumbs’, Slovenian); shịshị (Igbo). 
\\\hline 

1–N & 
Source idiom has multiple Target idioms &
\small
\underline{‘brownie points’} $\rightarrow$ jó pont or pirospont (Hungarian). 
\underline{‘brain drain’} $\rightarrow$ protų nutekėjimas; smegenų nutekėjimas (Lithuanian). 
\underline{‘blind alley’} $\rightarrow$ \textpersian{به ترکستان رفتن}/be Torkestan raftan (‘to go to Turkestan’, Farsi); \textpersian{آب در هاون کوبیدن}/ab dar havan kubidan (‘pounding water in a mortar’, Farsi). 
\underline{‘apples and oranges’} $\rightarrow$ \textarabic{بیرگازانا آتسان قینمز}/bir gazana atsan geynəməz (‘not to boil if placed together’, Azeri); \textarabic{الله لاری فرق المخ}/Allahları fərq eləməx (‘to have different Gods’, Azeri). 
\underline{‘cloud nine’} $\rightarrow$ \texthebrew{ברקיע השביעי}, \texthebrew{מאושר עד הגג} (Hebrew).\\\hline

N–1 & 
Multiple source idioms have same Target idiom &
\small
\underline{‘big fish’ and ‘big cheese’} $\rightarrow$ peix gros (Catalan); didelė žuvis (Lithuanian); veľké zviera (Slovak); \textarabic{بویوک باش}/böyük baş (Azeri). 
\underline{‘big cheese’, ‘big wig’, ‘big fish’} $\rightarrow$ важна клечка (‘vazhna klechka’, Bulgarian); stor kanon (Danish); \textpersian{کله گنده}/kalle gonde (Farsi); oke osisi (‘big tree’, Igbo). 
\underline{‘big wig’ and ‘big cheese’} $\rightarrow$ liels čiekurs (Latvian); большая шишка (Russian). 
\underline{‘blind alley’ and ‘brick wall’} $\rightarrow$ \texthebrew{מבוי סתום} (Hebrew). ‘eager beaver’ and ‘busy bee’ $\rightarrow$ vreden kao pčela (Serbian); vreden kao pčela/mravka (Macedonian). 
\underline{‘meat and potatoes’ and ‘bread and butter’} $\rightarrow$ arroz e feijão (Portuguese).
\\
\bottomrule
\end{tabular}
\end{adjustbox}
\caption{Cross-linguistic mappings of English idioms showing different types of equivalence and adaptation.}
\label{tab:idiom-mappings}
\end{table*}

The dataset includes data for \numLanguages{} language variants\footnote{Some languages in this list are still in the preparation stage, please see the forthcoming appendix for a precise status of each language.}, viz., Aromanian, Azeri, Bulgarian, Catalan, Chinese, Danish, Farsi, Georgian, Greek, Hebrew, Hungarian, Igbo, Indonesian, Italian, Javanese, Kazakh, Latvian, Lithuanian, Luxembourgish, Macedonian, Norwegian, Brazilian and European Portuguese, Russian, Serbian, Slovak, Slovenian, Ecuadorian and European Spanish, Swahili, Turkish, Ukrainian, Urdu, Uzbek with contributions from 89 language experts, producing over \numPIEtxt{} expressions from the English seed PIEs.
For a subset of those, 5 images were generated resulting in a total of \numPIEimg{} images. 

Across all languages, leaders reported structural, semantic, and cultural specificities (Table \ref{tab:idiom-mappings}). Although several idioms had an identical or near identical counterpart (1-1 in Table \ref{tab:idiom-mappings}), such as ``bad apple'', others implicated structural differences, as is the case of noun phrases such as ``close shave'' having as counterparts adverbial phrases, such as ``o vlások'' (`by a hair', Slovak). Others lacked direct lexical or idiomatic equivalents (1-0 in Table \ref{tab:idiom-mappings}), reflecting cultural or conceptual gaps that required paraphrasing or descriptive expressions. Even when equivalents existed, partial semantic mismatches were frequent, as figurative scope, tone, or emotional polarity rarely aligned perfectly across languages. Cultural imagery also played a role in finding suitable idiomatic equivalents, as metaphors grounded in local experience are often needed to replace or revise the original image generated for English. Some idioms became lexicalized into single words (1-0 in Table \ref{tab:idiom-mappings}), while others expanded into multiple variants or broader idioms with overlapping meanings (1-N in Table \ref{tab:idiom-mappings}). The influence of English was also reported through calques and loan translations.


Based on the  qualitative reports by the language experts,  distinct types of idiom mappings were identified, which characterise how idiomatic meanings differ across  languages, in relation to the English seed PIE,  as summarized in Table \ref{tab:idiom-mappings}.  
These  fine-grained mappings can be grouped into three rough-grained 
types of alignment  
with the English seed PIEs: 1-1, when the idiomatic expression is also realised idiomatically; 1-0, when the meaning is conveyed but not as an expression;  1-N, when multiple PIEs are available, possibly reflecting ambiguity or variation; and N-1, when multiple English PIEs are equivalent.  
Figure \ref{fig:translation_by_mwe} shows the proportion of a subset of types across English PIEs. Focusing on the English PIEs  which have an idiomatic parallel in most other languages, these include cases like  ``black market'' or ``cash cow''
which have correspondences across 90\% and 86.7\% of the languages, reflecting their conceptual salience and likely metaphorical shared concept as ``mercato nero'' in Italian (lit. ``market black''). On the other hand, several expressions (e.g., ``blue print'', ``baby blues'', ``box office'') often have no idiomatic equivalent, and instead their meaning is expressed through single words or non-idiomatic realisations.

\begin{figure}[htp!]
    \centering
    \begin{subfigure}{0.4\textwidth}
    \includegraphics[width=\linewidth]{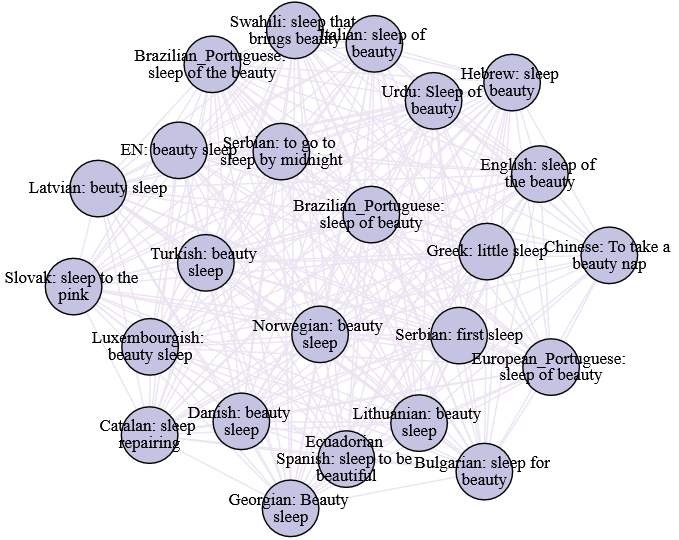}
        \caption{\emph{highly convergent} idiom (\textit{beauty sleep}). 
        A single large component with numerous connections among literal translations, indicating strong lexical transparency and widespread cross-linguistic lexicalisation.}
        \label{fig:beauty_sleep}
    \end{subfigure} 
    
    \begin{subfigure}{0.4\textwidth}
      \includegraphics[width=\linewidth]
      {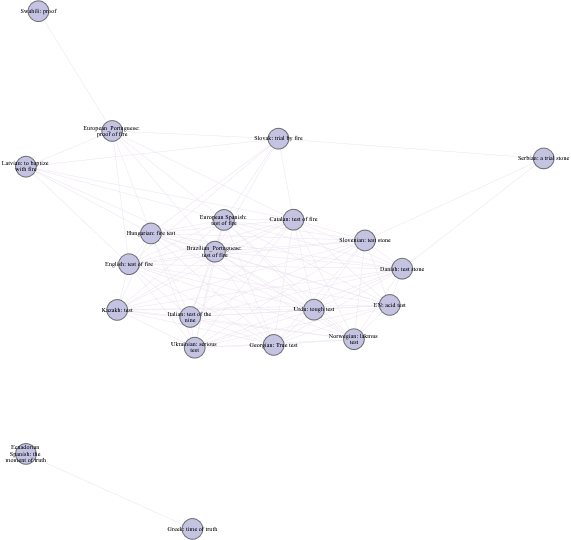} 
      \caption{\emph{divergent} example (\textit{acid test}). 
      One dominant component with smaller  clusters, 
      suggesting partial lexical convergence around a shared core metaphor.}
      \label{fig:acid_test}
    \end{subfigure}
\caption{Two representative idioms displaying distinct network structures.}
\label{fig:graph_examples}
\end{figure}

Taken together, the two perspectives reveal that cross-linguistic idiomatic equivalence is not evenly distributed. Some PIEs are conceptually stable and widely share a lexicalisation across languages, while others are culture-specific or structurally divergent. For example, ``beauty sleep'' is near-literally shared across numerous languages (Figure~\ref{fig:beauty_sleep}), whereas ``acid test'' exhibits a more fragmented pattern and tends to be rendered descriptively rather than idiomatically across languages (Figure~\ref{fig:acid_test}).
Moreover, their meanings cannot be straightforwardly transferred from one language to the other simply by translating their individual words. Instead some language and culture specific knowledge is required, even among variants of the same language. This is the case of  ``bacia sem fundo'' (lit. ``basin without bottom'') in Brazilian Portuguese and ``emplastro numa perna de pau'' (lit. ``plaster in a leg of wood'') in European Portuguese for ``chocolate teapot'' in English. 
This unevenness underscores the challenges of modelling idiomaticity in multilingual NLP and highlights the need for resources that explicitly capture when and how idiomatic meanings are lexicalized differently across languages.


To capture patterns of cross-linguistic similarity among idiomatic expressions, we also represented each English PIE and its equivalents as graphs. For each English PIE, a graph was generated where each node corresponds to an equivalent language-specific PIE (represented in the graph via its literal translation to English), and where the edges represent lexical overlap between the PIEs in  different languages. This means that two nodes are connected when, after stopword removal, they share at least one content word. 

This heuristic can generate more than one subgraph for a given PIE, reflecting clusters of common lexicalisations. For any PIE, the number of nodes in the largest connected component of the graph reveals how many languages share, at least in part, lexicalisations. This means that a single-component graph would indicate that all languages share part of the PIE lexicalisation, while multiple sub graphs for a given PIE  represent clusters of lexically related PIEs. 
Indeed, these graphs can highlight cross-lingual lexical overlaps between PIE equivalents in different languages, and Figure \ref{fig:graph_examples} shows the graphs for two representative PIEs. For example,  ``beauty sleep'' is lexicalized using the multilingual equivalents of these words in a large variety of languages (Figure \ref{fig:graph_examples}a). On the other hand, the resulting graphs may be disconnected, depending on how varied the lexicalisations of the idiomatic elements are across languages, as is the case for ``acid test'' (Figure \ref{fig:graph_examples}b). 

For each PIE, we computed structural graph measures that capture the overall connectivity and lexical cohesion across languages (see Table~\ref{tab:graph_measures}). These include the number of nodes and edges, the number of connected components, the proportion of nodes contained in the largest component, and the overall graph density. Edge weights quantify the amount of shared lexical material between translations, while the vocabulary size reflects the total number of distinct content words used. Finally, clique-based metrics identify subsets of languages whose literal translations are mutually related, providing a complementary view of complete lexical convergence.

\if{}
\begin{table*}[htp!]
\centering
\small
\begin{tabular}{llllll}
\hline
\textbf{PIE-EN} & \textbf{LITERAL} & \textbf{LITERAL} & \textbf{IDIOM} & \textbf{IDIOM} & \textbf{IDIOM} \\
 & \textbf{Translation} & \textbf{Transliteration} & \textbf{Equivalent} & \textbf{Translation} & \textbf{Transliteration} \\
\hline

\hline
bad apple  & ruim maça & - &  maça podre & apple rotten & - \\

bad apple  & gnile âbluko & 
    гниле яблуко  &  paršiva vìvcâ & lousy sheep & паршива вівця\\ 

black box	& caixa preta & - & caixa-preta & black-box & \\
beauty sleep	& scheinheet schloof & - & Schéinheetsschloof & beauty sleep & - 
\\\hline
\end{tabular}
\caption{Annotation Schema - Examples
}
\label{tab:annotation}
\end{table*}
\todo[inline]{RW: AV you asked for some examples Tab.\ref{tab:annotation} but I think we should drop Tab \ref{tab:annotation} and keep Tab 2}

\fi

\begin{table*}[htp!]
\centering
\footnotesize
\begin{tabular}{lrrrrl}
\hline
\textbf{Metric} & \textbf{Mean} & \textbf{SD} & \textbf{Min} & \textbf{Max} & \textbf{Interpretation} \\
\hline
Nodes per PIE graph & 12.0 & 8.9 & 4 & 53 & Number of languages represented per idiom \\
Edges per PIE graph & 37.2 & 77.4 & 1 & 18 & Lexical connections between PIE equivalents \\
Number of components & 5.0 & 3.1 & 1 & 18 & Fragmentation of idiom networks \\
Largest component ratio & 0.56 & 0.22 & 0.11 & 1.00 & Proportion of nodes in the main cluster \\
Density & 0.31 & 0.25 & 0.01 & 1.00 & Overall connectivity among PIEs \\
Mean edge weight & 1.55 & 0.46 & 1.0 & 3.33 & Average number of shared words \\
Vocabulary size (unique words) & 12.4 & 10.0 & 2 & 60 & Lexical diversity in PIEs \\
Clique number & 6.11 & 5.76 & 2 & 31 & Size of the largest fully connected component \\
Number of maximal cliques & 5.76 & 4.0 & 1 & 21 & Number of fully connected subgraphs  \\
\hline
\end{tabular}
\caption{Summary statistics of graph-based measures across idioms.}
\label{tab:graph_measures}
\end{table*}

Across the PIEs analysed, the graphs exhibit considerable structural variability. Most PIEs form fragmented networks (on average eight components per graph), indicating limited cross-linguistic lexical overlap. Only a few PIEs, such as ``beauty sleep'' and ``black box'', display dense, single-component structures, reflecting high transparency and widespread lexicalisation.

The average density (0.31) confirms that idioms generally form partially connected clusters rather than a fully interconnected graph with a single lexicalisation across languages. The largest component typically contains about 60\% of the nodes, showing that some PIEs maintain a common lexical core shared across many languages, while others are divided into several smaller, internally consistent groups. Clique metrics reinforce this distinction: most PIEs contain numerous small cliques, while only a few reach large clique numbers, corresponding to PIEs whose lexicalisation  is shared across languages. Such groupings may reflect genealogical proximity or shared cultural norms, but the analysis remains agnostic as to their underlying causes. Two representative scenarios with  distinct network structures are displayed in 
Figure~\ref{fig:graph_examples}.

\section{Multilingual and Multimodal Idiomatic Representations}
\label{sec:results}

We analyze the XMPIE dataset and evaluate a publicly available vision--language baseline model, to showcase the challenges for language and vision models related to accurate idiomatic understanding. Among the challenges that this dataset can probe are those related to the   multilingual idiomatic representation abilities of a model. In particular, the question of to what extent  idiomatic  representation is shared across languages, and if understanding in one language can lead to understanding in other languages. There are also questions related to multiple modalities, and to what degree accurate representation in one modality (like text) is also shared across modalities (like vision).

\begin{table}[htp!]
  \centering
  \footnotesize
  \begin{tabular}{lccccc}
    \toprule
    \textbf{Lg} & \textbf{T1-I} $\uparrow$ & \textbf{T1-L} $\uparrow$ & \textbf{T2-I} $\uparrow$ & \textbf{T2-L} $\uparrow$ & \textbf{NDCG@5} $\uparrow$ \\
    \midrule
    EN & 0.060 & 0.900 & 0.010 & 0.660 & 0.954 \\
    BP & 0.100 & 0.580 & 0.020 & 0.360 & 0.910 \\
    ES & 0.375 & 0.234 & 0.125 & 0.094 & 0.896 \\
    CN & 0.140 & 0.316 & 0.018 & 0.140 & 0.877 \\
    TR & 0.143 & 0.232 & 0.000 & 0.018 & 0.876 \\

    \bottomrule
  \end{tabular}
  \caption{PIE-only, no-training results with \textbf{EVA-CLIP-18B}. Top-2 (T2) is strict (both ranks must match). NDCG@5 uses gains $(1,0.5,0.5,1,0)$.}
  \label{tab:eva18b}
\end{table}

\subsection{Experimental Setup}
\label{sec:setup}
We evaluate systems on how well they rank five candidate images for each PIE. Each item provides five standardized image slots: one \textbf{idiomatic} image (image 1), one \textbf{literal} image (image 4), two \textbf{weak variants} (images 2–3), and one \textbf{distractor} (image 5).

This evaluation helps to determine if models are able to identify either the literal or the idiomatic target senses among the five images.  
This is an initial evaluation without sentences to provide contextual clues that could help to disambiguate between idiomatic and literal senses. Scores are computed per item and averaged per language.

\subsection{Evaluation Metrics}
\label{sec:evaluation}
\paragraph{Top-1 Accuracy (T1):}
We report two Top-1 settings: (i) \textbf{Idiomatic Top-1} (T1-I, correct if rank~1 is the idiomatic image); (ii) \textbf{Literal Top-1} (T1-L, correct if rank~1 is the literal image).
\vspace{-0.3cm}
\paragraph{Top-2  Accuracy (T2):}
This metric is more strict and requires that both rank~1 and rank~2 correspond to the idiomatic and literal senses (or vice-versa).
\vspace{-0.3cm}
\paragraph{Normalised Discounted Cumulative Gain (NDCG@5):}
To reflect graded usefulness across the whole list, we use NDCG@5. We assign gains based on the image slot:
\[
g=(1,\;0.5,\;0.5,\;1,\;0)\quad\text{for slots }(1\ldots5).
\]
This means the idiomatic (image 1) and literal (image 4) options receive equal credit (of 1), the two weak variants (images 2–3) receive partial credit (0.5), and the distractor (image 5) receives no credit (0).

Given a system ranking $\pi$, where $\pi(i)$ denotes the image slot placed at rank $i$, the relevance at position $i$ is defined as $rel_i = g_{\pi(i)}$. \citetlanguageresource{pickard-etal-2025-semeval} report DCG; here we use the \emph{normalised} variant to enable comparability across items.
Let $rel_i$ be the gain at rank $i$ and $rel_i^{(\text{ideal})}$ the gain at rank $i$ in the ideal ordering. Then
\[
\mathrm{NDCG}@5 \;=\;
\frac{\displaystyle \sum_{i=1}^{5} \frac{rel_i}{\log_2(i+1)}}
{\displaystyle \sum_{i=1}^{5} \frac{rel_i^{(\text{ideal})}}{\log_2(i+1)}} \in [0,1].
\]

\subsection{Baseline Method}
We adopt \textbf{EVA-CLIP-18B} \citeplanguageresource{sun2024eva} as a no training, retrieval-style baseline, building on CLIP \citeplanguageresource{radford2021learning} because its joint image-text space enables simple, reproducible cosine-similarity ranking, and the large-scale EVA-CLIP-18B variant offers excellent retrieval performance. For each item (PIE $c$), the PIE itself is used as the only text query and is compared against five candidate images in the model's joint embedding space.
\vspace{-0.3cm}
\paragraph{Query formulation:}
The text input is simply the PIE string $c$, with no prompt templating or additional context (e.g., ``beauty sleep''). A single text embedding is computed for $c$ and scored against the five images of the item.
\vspace{-0.3cm}
\paragraph{Scoring and ranking:}
Let $t=f_{\text{text}}(c)$ and $i_j=f_{\text{img}}(I_j)$ be EVA-CLIP-18B text and image embeddings. We L2-normalize and compute cosine scores
\[
s_j=\hat i_j^\top \hat t,
\]
then rank images by $s_j$ (descending); the top-ranked image is the prediction.

\subsection{Results}

We report PIE-only results for five languages with off-the-shelf pre-trained models without additional training or fine-tuning: English (EN), Brazilian Portuguese (BP), Ecuadorian Spanish (ES), Chinese (CN), and Turkish (TR). These languages were selected to represent a diverse language sample. For these we show \emph{Idiomatic/Literal} Top-1, \emph{Idiomatic/Literal} Top-2 accuracy, and the average NDCG@5 with symmetric gains $(1,0.5,0.5,1,0)$.
NDCG@5 provides a single interpretable summary of ranking quality, while Top-1 preserves an intuitive success rate.


As shown in Table~\ref{tab:eva18b}, \textbf{NDCG@5} is consistently high across languages (EN~0.954, BP~0.910, ES~0.896, CN~0.877, TR~0.876), indicating that—even without context—the model typically ranks the target and weak variants near the top. However, \textbf{Top-1} and especially the stricter \textbf{Top-2} reveal a clear targeting asymmetry: for EN/BP/CN/TR, the models consistently rank the \emph{literal} image consistently,  substantially outperforming the ranking of the  \emph{idiomatic} image (e.g., EN Top-1: 0.900 vs.\ 0.060; Top-2: 0.660 vs.\ 0.010). However, ES is the notable exception (idiomatic $>$ literal both on Top-1 and Top-2). The low idiomatic Top-2 values show that getting \emph{both} the idiomatic image and its semantically-related distractor into the top two \emph{in the correct order} is much harder than simply placing relevant images near the top (as NDCG suggests).

\emph{Examples:} We observe several NDCG-perfect cases under the weak-half gains:
EN---\enquote{pipe dream}, \enquote{ghost town}, \enquote{watering hole}, \enquote{flying saucer};
BP---\enquote{colocar a boca no trombone}, \enquote{mercado de pulgas};
ES---\enquote{vacaciones sin descanso}, \enquote{sinverg\"uenza}, \enquote{premio acad\'emico};
CN---\enquote{\textchinese{黑箱}};
TR---\enquote{b\"uy\"uk resim}.
Despite these strong cases, strict Top-2 remains low for idiomatic senses in most languages, confirming that fine-grained idiomatic disambiguation is still a bottleneck in model performance. Although in this evaluation only the PIE is provided, without any contextual clues to aid interpretation, the more relaxed evaluation measures adopted take this into account by allowing any ordering of the target senses in the first positions as acceptable.

\section{Conclusion}
\label{sec:conclusion}

We presented XMPIE, a multilingual and multimodal idiomaticity dataset covering around 10K items for \numLanguages{} language variants.
This parallel resource 
allows cross-lingual analyses about the realisation of different concepts idiomatically providing insights into the salient linguistic and cultural aspects 
while also enabling assessment of the multilingual abilities of models and the extent to which understanding of an idiomatic expression transfers across languages and modalities.\footnote{XMPIE is available on ANONYMOUS link which will be made publicly available upon acceptance.} The results obtained with a baseline model on a subset of the dataset,
reveal substantial cross-lingual variation and a consistent literal-over-idiomatic advantage in PIE-only experiments, underscoring the need for contextual cues for robust idiom understanding. Future work includes extending XMPIE 
and adding information about factors that may play a role in model and human processing like abstractness and imageability.         

\section*{Acknowledgments}
This work received support from the CA21167 COST action UniDive, funded by COST (European Cooperation in Science and Technology).




\section*{Ethical considerations and limitations}

\paragraph{Item vetting:}
During content creation, annotators could (i) mark items as successfully generated; (ii) flag problematic items with notes; or (iii) propose removal. Removal reasons included: always-literal expressions, obsolete items, offensive content, or cases where a viable literal rendering was unattainable. 

\paragraph{Safety filters:}
Guidelines explicitly avoided swearing, illegal activity, and negatively framed mentions of specific people/organisations in context sentences; image prompts similarly steered clear of upsetting or harmful content. 

\paragraph{Cross-lingual Effects:}
Because for nuanced language image generation models seem to perform best with English prompts, contributors translated non-English prompts where helpful, while preserving language-specific idiomatic content in the contexts and item selection. 
Therefore there may be nuanced effects that are not represented given the cultural language bias in the image generation models. 

\nocite{*}
\section{Bibliographical References}\label{sec:reference}
\renewcommand{\refname}{}
\bibliographystyle{lrec2026-natbib}

\bibliography{lrec2026-example}

\section{Language Resource References}

\bibliographystylelanguageresource{lrec2026-natbib}
\bibliographylanguageresource{languageresource}

\end{document}